\documentclass[letterpaper]{article} 
\usepackage{aaai2027}  
\usepackage[hyphens]{url}  
\usepackage{graphicx} 
\usepackage{natbib}  
\usepackage{caption} 
\usepackage{algorithm}
\usepackage{algorithmic}

\usepackage{newfloat}
\usepackage{listings}
\DeclareCaptionStyle{ruled}{labelfont=normalfont,labelsep=colon,strut=off} 
\floatstyle{ruled}
\newfloat{listing}{tb}{lst}{}
\floatname{listing}{Listing}

\usepackage{booktabs}
\usepackage{amsmath}
\usepackage[table]{xcolor}
\usepackage{pifont}

\definecolor{ourshade}{RGB}{255,244,232}
\definecolor{grassgreen}{RGB}{137,191,61}

\title{V-STORM: A Structure-Aware Grounded Chain of Thought Dataset for Faithful and Fine-Grained Spatio-Temporal Video Reasoning}
\author{
    Written by AAAI Press Staff\textsuperscript{\rm 1}\thanks{With help from the AAAI Publications Committee.}\\
    AAAI Style Contributions by Peter Patel Schneider,
    Sunil Issar,\\
    J. Scott Penberthy,
    George Ferguson,
    Hans Guesgen,
    Francisco Cruz\equalcontrib\corresponding,
    Marc Pujol-Gonzalez\equalcontrib\corresponding
}
\affiliations{
    \textsuperscript{\rm 1}Association for the Advancement of Artificial Intelligence\\

    1101 Pennsylvania Ave, NW Suite 300\\
    Washington, DC 20004 USA\\
    proceedings-questions@aaai.org
}

\title{V-STORM: A Structure-Aware Grounded Chain of Thought Dataset for Faithful and Fine-Grained Spatio-Temporal Video Reasoning}
\author {
    Dian Jiao\textsuperscript{\rm 1}\equalcontrib,
    Yuqian Yuan\textsuperscript{\rm 1}\equalcontrib,
    Jiaqi Zhu\textsuperscript{\rm 2},
    Mingjian Gao\textsuperscript{\rm 1},
    Siwen Tan\textsuperscript{\rm 1},
    Yunqi Cao\textsuperscript{\rm 1},
    Jinhui Li\textsuperscript{\rm 1},
    Sijing Li\textsuperscript{\rm 1},
    Zhenxuan Fan\textsuperscript{\rm 1},
    Haoyang Wang\textsuperscript{\rm 1},
    Wenqiao Zhang\textsuperscript{\rm 1}\corresponding,
    Siliang Tang\textsuperscript{\rm 1}\corresponding,
}
\affiliations {
    \textsuperscript{\rm 1}Zhejiang University\\
    \textsuperscript{\rm 2}National University of Singapore\\
    jd\_dcd@zju.edu.cn
}

\begin{document}
\nocopyright
\maketitle

\begin{abstract}
While MLLMs can often identify \textbf{\emph{what}} happens in a video, they still struggle with
fine-grained spatio-temporal reasoning, which requires precise grounding \textbf{\emph{when}} and \textbf{\emph{where}} the supporting evidence appears and composing such evidence into a coherent reasoning chain. This challenge is exacerbated by the lack of scalable data construction and evaluation pipelines that provide granular, verifiable supervision linking reasoning steps to explicit spatio-temporal evidence.
To address this gap, we present \textbf{\texttt{V-STORM}}, a structure-aware framework that transforms videos into Spatio-Temporal Scene Graphs (STSGs) and derives Grounded Chain of Thought (GCoT) from validated evidence paths, making each reasoning step traceable to structured and verifiable spatio-temporal evidence.
Under this framework, we construct a 260K-sample training set and a 3K-sample human-verified benchmark, both featuring GCoT annotations that link reasoning to explicit spatio-temporal evidence.
Experiments reveal a clear separation between evidence localization and evidence composition: current MLLMs can often identify relevant spatio-temporal cues, yet still fail to integrate them into correct multi-step reasoning, exposing a bottleneck that answer-only evaluation largely obscures. Moreover, fine-tuning with V-STORM substantially improves both answer accuracy and grounding faithfulness, validating its effectiveness as supervision for faithful and evidence-grounded video reasoning.
Our dataset, benchmark, and source code will be publicly released.
\end{abstract}


\begin{figure*}
  \centering
  \includegraphics[width=0.94\textwidth]{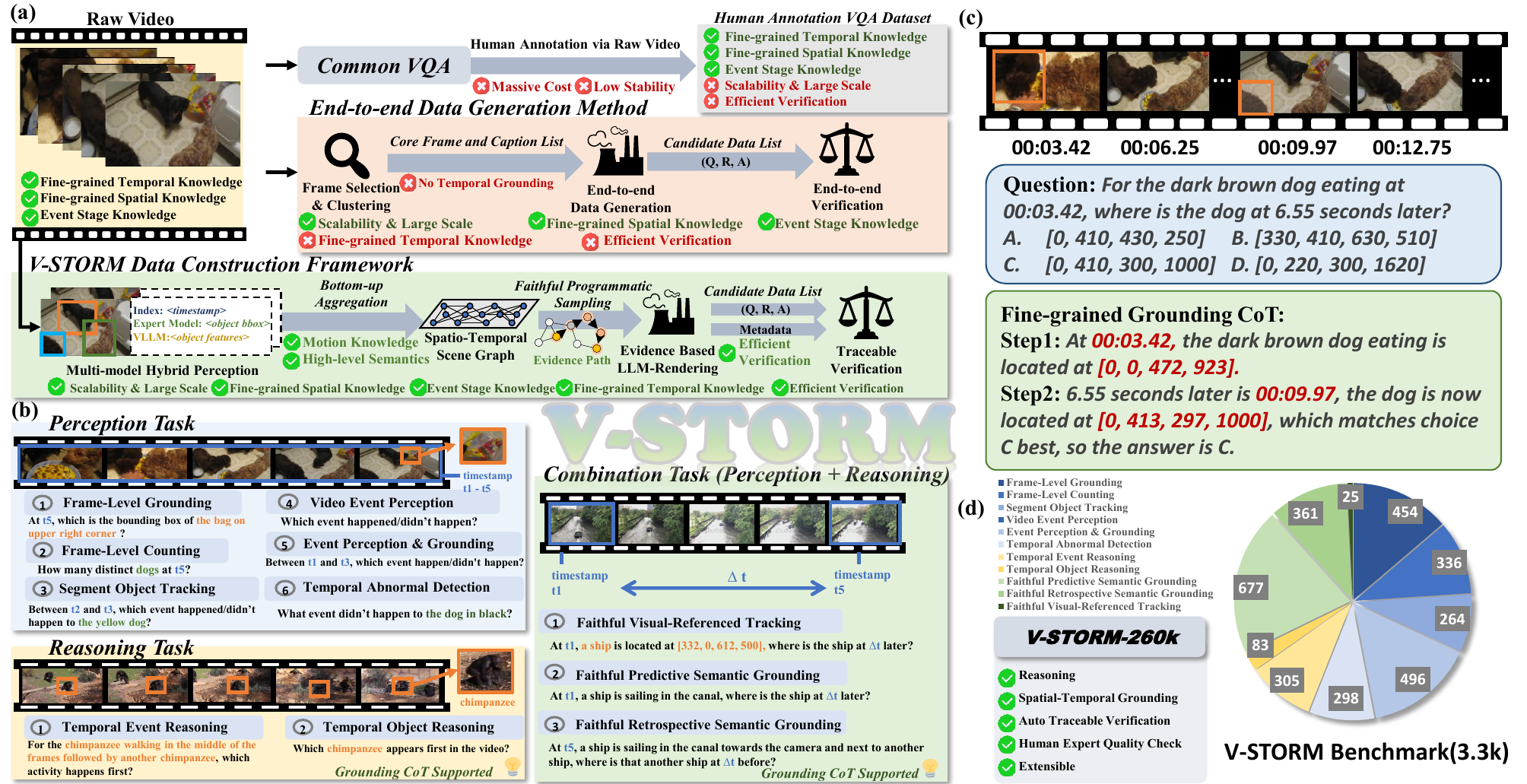} 
  \caption{
    \textbf{Overview of V-STORM.}
    \textbf{(a) Construction Pipeline:} Unlike traditional methods, V-STORM employs a \textit{structure-aware framework} with multi-model collaboration to generate complex tasks and verifiable CoT annotations.
    \textbf{(b) Task Taxonomy:} We design \textbf{11 tasks} spanning \textit{Perception}, \textit{Reasoning}, and \textit{Combination} dimensions.
    \textbf{(c) Examples:} Sample queries demonstrating fine-grained grounding (specific timestamps and coordinates) in both questions and GCoT.
    \textbf{(d) Statistics:} Dataset distribution highlighting the scale (260k), diversity, and high-quality verification.
  }
  \label{fig:intro}
\end{figure*}


\maketitle

\begin{table*}[t]
\centering
\caption{\textbf{Comparison of representative video reasoning datasets.}
We compare existing datasets across scale, grounding, reasoning capability, and construction paradigm.
\textbf{Compositional ST-Reasoning} denotes multi-hop spatio-temporal reasoning ability.
\textbf{GCoT} uses language-based spatio-temporal reasoning traces,
while \textbf{MMCoT} denotes interleaved multimodal CoT.}

\label{tab:comparison}

\small
\setlength{\tabcolsep}{1mm}
\renewcommand{\arraystretch}{1.0}

\begin{tabular}{l |c |cc c cc |c c}
\toprule
\textbf{Dataset} 
& \textbf{Training Set} 
& \multicolumn{2}{c}{\textbf{Grounding}} 
& \textbf{Compositional} 
& \multicolumn{2}{c}{\textbf{Annotation}} 
& \textbf{Construction} 
& \textbf{Video-Free} \\

\cmidrule(lr){3-4} \cmidrule(lr){6-7}

& \textbf{Scale}
& \textbf{Spatial} 
& \textbf{Temporal} 
& \textbf{ST-Reasoning} 
& \textbf{CoT} 
& \textbf{Verifiable} 
& \textbf{Paradigm}
& \textbf{Extension}\\

\midrule
\midrule
Video-MME\cite{fu2025video}      
& -- & -- & -- & -- & -- & -- & Human & -- \\

MotionBench\cite{hong2025motionbench}    
& -- & -- & -- & -- & -- & -- & Human & -- \\

NExT-QA\cite{xiao2021next}        
& -- & -- & -- & -- & -- & -- & Human & -- \\

ECBench\cite{dang2025ecbench}        
& -- & -- & -- & -- & -- & -- & Human & -- \\

VCR-Bench\cite{qi2025vcr}       
& -- & \ding{51}  & \ding{51}  & -- & GCoT & -- & LLM+Human & -- \\

\midrule

E.T. Bench\cite{liu2024bench}     
& -- & -- & \ding{51}  & -- & -- & -- & Human & -- \\

EOC-Bench\cite{yuan2025eoc}     
& -- & \ding{51}  & -- & -- & -- & -- & Human & -- \\

\midrule

Video-Espresso\cite{han2025videoespresso} 
& 203k & \ding{51}  & -- & -- & MMCoT & -- & LLM & -- \\

\rowcolor{ourshade}
\textbf{V-STORM (Ours)} 
& \textbf{260k} 
& \textbf{\ding{51} } 
& \textbf{\ding{51} } 
& \textbf{\ding{51} } 
& \textbf{GCoT} 
& \textbf{\ding{51} } 
& \textbf{Programmatic} 
& \textbf{\ding{51} } \\

\bottomrule
\end{tabular}
\end{table*}

\section{Introduction}
\label{sec:intro}

Recent advances in multi-modal large language models (MLLMs) have significantly improved video-level understanding and question answering~\citep{bai2025qwen2, liu2023visual, hurst2024gpt}. However, these models still struggle with \emph{fine-grained spatio-temporal reasoning}. This capability requires not only answering \emph{what} happens in a video, but also explicitly grounding \emph{when} and \emph{where} the supporting evidence lies and correctly integrating such evidence into a coherent reasoning process. While recent studies have begun to enhance fine-grained temporal~\citep{han2025videoespresso} or spatial~\citep{yuan2025eoc} understanding, and emerging works explore evidence-grounded reasoning~\citep{wang2025traceable, feng2025video, wang2025videorft, meng2026openo3videogroundedvideoreasoning}, these efforts are often limited to isolated aspects or model-level designs, lacking systematic support for jointly modeling spatio-temporal grounding with structured, verifiable reasoning.

Fine-grained spatio-temporal reasoning involves three interdependent capabilities: (i) temporal grounding identifying when relevant events occur; (ii) spatial grounding localizing the involved entities; and (iii) compositional grounding integrating grounded observations into multi-step chains. These capabilities must be evaluated separately: accurate localization does not guarantee correct reasoning. In high-stakes settings like surveillance or embodied intelligence, plausible answers are insufficient without grounding in the right evidence. For example, a model may correctly report that a person entered a restricted area, but ground it to the wrong time window or individual---making the decision unreliable and hard to audit.

A major bottleneck is the lack of scalable data that jointly provides fine-grained grounding, explicit reasoning supervision, and verifiable evidence.
As Table~\ref{tab:comparison} summarizes, most existing video reasoning datasets suit evaluation but offer no step-level supervision for training evidence-grounded models~\citep{fu2025video, hong2025motionbench, xiao2021next, dang2025ecbench, qi2025vcr}.
A few~\citep{liu2024bench, yuan2025eoc} provide high-quality spatial or temporal annotations, yet rely on costly expert labeling and offer limited reasoning depth.
Video-Espresso~\citep{han2025videoespresso} scales data via LLM generation, but focuses on high-level reasoning without fine-grained spatio-temporal grounding or verifiable supervision.
Thus, existing resources insufficiently support faithful fine-grained spatio-temporal reasoning, where models must both locate and compose evidence into multi-step chains.

To address these limitations, we propose \textbf{\texttt{V-STORM}} (Fig.~\ref{fig:intro}), a structure-aware framework formulating video reasoning as a graph-grounded process for scalable and evidence-aligned data construction.
The key component of V-STORM is the Spatio-Temporal Scene Graph (STSG), which organizes objects, trajectories, and dynamic relations into a unified and reusable representation. 
Unlike generation-based approaches that directly synthesize reasoning chains, STSG provides an explicit intermediate structure that preserves fine-grained spatial and temporal evidence. 
Built from complementary perception and MLLMs, a single STSG supports diverse reasoning paths and downstream tasks without repeatedly processing the original video. 
Based on this representation, V-STORM performs graph-guided question generation, where QA pairs and Grounded Chain-of-Thought (GCoT) annotations are derived from valid evidence paths. 
Consequently, each reasoning step is associated with explicit temporal and spatial evidence, enabling more reliable supervision and systematic evaluation.

Based on this framework, we construct the V-STORM Dataset with over 260K training samples and a 3.3K human-verified benchmark. The benchmark introduces a grounding-aware evaluation protocol that separately assesses perception, evidence localization, and compositional reasoning, revealing distinct failure modes of current MLLMs.

Our contributions are summarized as follows:
\begin{itemize}
    \item We propose \textbf{V-STORM}, a structure-aware framework that generates scalable and evidence-grounded video reasoning supervision through graph-based modeling.
    
    \item We introduce the \textbf{V-STORM Dataset}, containing \textbf{260K} training samples and a \textbf{3.3K} human-verified benchmark with Grounded Chain-of-Thought annotations for fine-grained spatio-temporal reasoning.
    
    \item We develop a grounding-aware evaluation protocol that disentangles perception, evidence localization, and reasoning, enabling systematic analysis of MLLM capabilities and limitations.
    
    \item Experiments show that current MLLMs can effectively localize evidence for relatively simple reasoning, but struggle with composing multiple grounded observations. V-STORM provides effective structured supervision improving both reasoning accuracy and evidence grounding.
\end{itemize}

\section{Related Work}

\noindent\textbf{Video Datasets.}
Existing video datasets generally fall into two categories. Benchmark-oriented datasets \cite{fu2025video, hong2025motionbench, xiao2021next, dang2025ecbench} focus on high-level video understanding with coarse video-level supervision, while localization-centric datasets \cite{han2025videoespresso, yuan2025eoc, liu2024bench} provide fine-grained temporal or spatial annotations but typically support limited reasoning complexity. Recent efforts enrich annotations or introduce more challenging reasoning tasks, yet no existing benchmark jointly supports structured reasoning with verifiable fine-grained spatio-temporal grounding.

\noindent\textbf{Video-LLMs.}
Recent Video-LLMs, including the Qwen-VL series \cite{wang2024qwen2,bai2025qwen2,bai2025qwen3} and InternVL series \cite{chen2024internvl,chen2024expanding,wang2025internvl3}, have substantially improved video understanding through stronger visual encoders, longer context modeling, and video-native architectures. Inference-time approaches such as Chain-of-Frames \cite{ghazanfari2025chain} and Temporal-CoT \cite{zhang2025video} further enhance temporal reasoning by explicitly organizing intermediate reasoning steps over video frames.

\noindent\textbf{Grounded Reasoning and Thinking-with-Videos.}
Recent work stresses traceable reasoning. TreeBench \cite{wang2025traceable} shows the need to assess reasoning--evidence alignment, while recent video-reasoning methods \cite{meng2026openo3videogroundedvideoreasoning, wang2025videorft, feng2025video} embed spatio-temporal evidence in reasoning. Yet these efforts target model design or training, and lack datasets providing reasoning traces, grounding labels, and sound metrics.

\begin{figure*}[ht]
  \centering
  \includegraphics[width=0.98\textwidth]{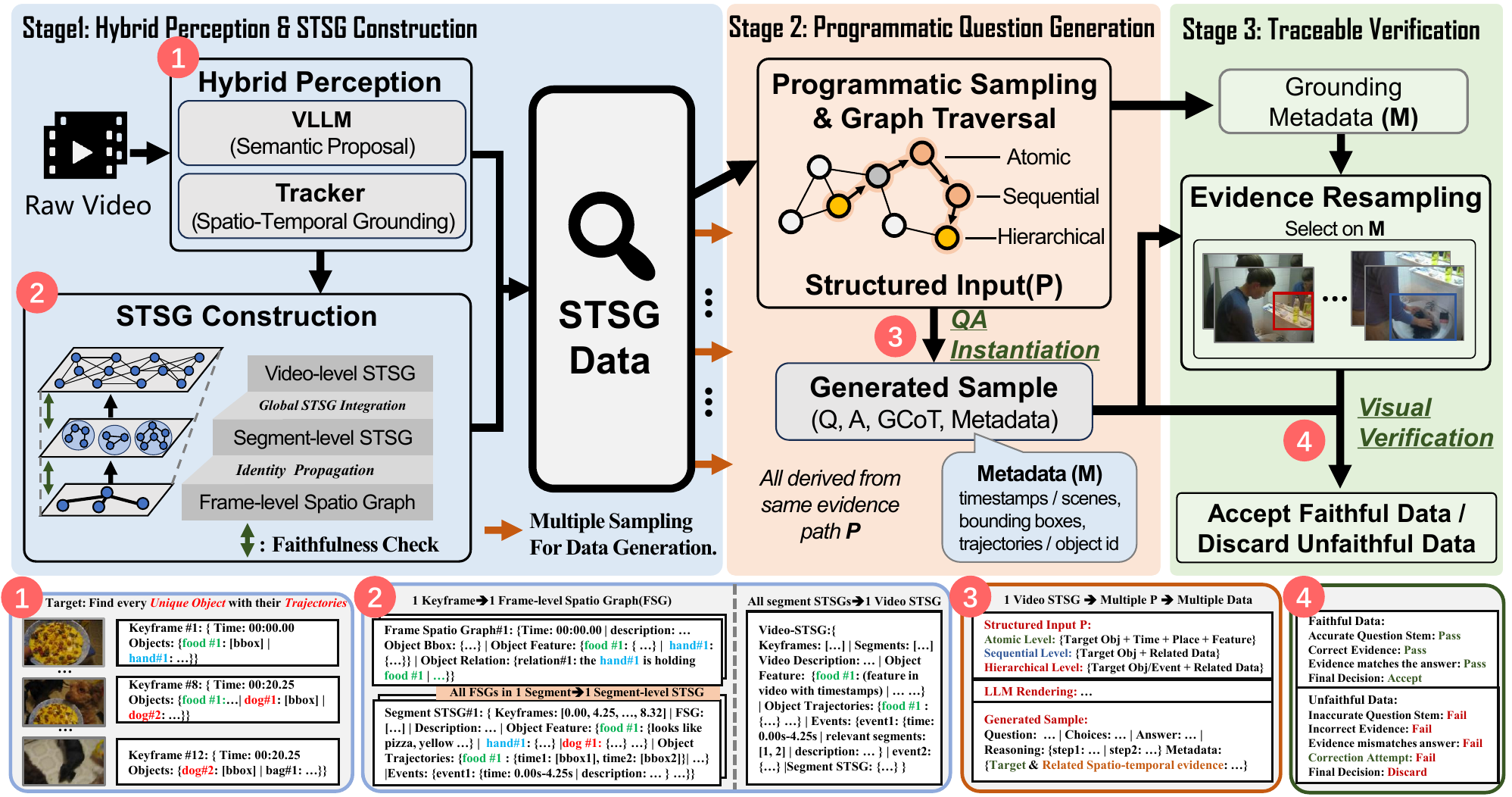} 
  \caption{\textbf{V-STORM Data Construction Pipeline.} The pipeline consists of three stages: (i) Hybrid Perception \& STSG Construction, (ii) Programmatic Question Generation, and (iii) Traceable Verification. The STSG serves as the core asset that connects raw video to generated structured samples.}
  \label{fig:pipeline}
\end{figure*}

\section{The V-STORM Framework}
V-STORM formulates fine-grained video reasoning as a graph-grounded data construction process. Questions, answers, and CoT annotations are derived from explicit spatio-temporal evidence rather than generated directly from raw videos or captions.
Based on videos, V-STORM generates question-answer pairs together with grounded Chain-of-Thought (GCoT) and corresponding spatio-temporal evidence annotations, enabling both reasoning supervision and evidence verification.

The framework consists of three stages: (i) \textbf{STSG Construction} transforms raw videos into structured spatio-temporal evidence; (ii) \textbf{Programmatic Generation} samples evidence paths from the STSG and renders them into question-answer pairs with GCoT annotations; and (iii) \textbf{Traceable Verification} validates the generated samples against the underlying evidence structure. Ultimately, V-STORM reduces the risk of ungrounded reasoning chains and enables systematic evaluation of both evidence localization and reasoning composition. The complete formal problem formulation is provided in the Supplementary Material, Sec. 2. Furthermore, to facilitate review and future reproducibility, we provide the anonymized source code in the Supplementary Material source code package, along with detailed descriptions of each file's functionality in Supplementary Material Sec. 3.

\subsection{Hybrid Perception \& STSG Construction}

Fine-grained spatio-temporal reasoning requires both accurate visual grounding and reliable semantic abstraction. Existing methods typically specialize in only one aspect: trackers such as SAM2 provide stable localization but limited semantics, while VLLMs enable open-vocabulary understanding yet often suffer from unreliable long-term grounding. To leverage their complementary strengths, we propose a hierarchical graph construction pipeline that separates low-level perception from high-level semantic abstraction, progressively organizing visual information from frame-level observations to segment-level structures and finally a unified video-level STSG.

\noindent\textbf{Hybrid Perception.}
Input video $\mathcal{V}$ is divided into temporal segments $\mathcal{S}=\{S_k\}_{k=1}^{K}$.
Within each segment, a semantic proposer (Florence-2) detects open-vocabulary object candidates on keyframes, while SAM2 propagates them into temporally consistent trajectories.
An MLLM is further employed to resolve ambiguous identities when it cannot be reliably determined from low-level visual cues.
Trajectories with insufficient temporal support are discarded, yielding a set of verified spatio-temporal elements $\{\tau_i\}$.

\noindent\textbf{Frame-level STSG Instantiation.}
Verified trajectories are instantiated as frame scene graphs
$\mathcal{G}_{\text{FSG}}^{(t)}=(\mathcal{V}_t,\mathcal{E}_t)$,
where each node represents an object with grounded attributes and each edge denotes a visually verified spatial relation.
This atomic representation preserves faithful frame-level evidence without introducing global abstraction.

\noindent\textbf{Hierarchical STSG Aggregation.}
Frame scene graphs are first aggregated within each temporal segment to form Segment-STSGs by propagating identities only inside short temporal windows, capturing object evolution while avoiding unreliable long-range association.
The resulting Segment-STSGs are then conservatively merged into a global Video-STSG through high-confidence cross-segment matching.
Finally, we build a reference index that links high-level events to their underlying atomic observations, enabling all subsequent question generation and verification to be traced back to validated visual evidence.

\subsection{Programmatic Question Generation \& Verification}

Existing video question answering (QA) datasets typically rely on end-to-end generation, which often produces visually ungrounded questions and unverifiable reasoning chains. In contrast, we generate each sample directly from the STSG, ensuring that questions, reasoning steps, and answers are explicitly grounded in spatio-temporal evidence rather than free-form language priors.

As illustrated in Fig.~\ref{fig:pipeline}, our generation process follows three stages: (i) \textbf{Programmatic Sampling} extracts a structured reasoning path from the STSG; (ii) \textbf{QA Instantiation} converts the sampled path into a QA pair with reasoning and metadata; and (iii) \textbf{Traceable Verification} checks faithfulness against visual evidence and filters inconsistent samples.

\noindent\textbf{Evidence Sampling.}
We first sample an evidence path $\mathcal{P}$ from the STSG $\mathcal{G}$. Each path consists of a sequence of connected objects, events, and relations with explicit spatial and temporal evidence. As shown in Stage 2 of Fig.~\ref{fig:pipeline}, the sampled paths can represent different reasoning types, including (i) atomic facts, (ii) temporal event sequences, and (iii) hierarchical event relations. This allows the same sampling strategy to support both simple perception tasks and more complex reasoning tasks. To construct challenging multiple-choice questions, we also sample visually plausible but incorrect candidates from the STSG as distractors.

\noindent\textbf{QA Instantiation.}
Given a sampled evidence path $\mathcal{P}$, we employ an LLM as a structure-preserving translator to generate a Question--Reasoning--Answer triplet together with grounding metadata $\mathcal{M}$ and distractors $\mathcal{D}$, yielding $(Q,R,A,\mathcal{M},\mathcal{D})$. The translator is constrained by $\mathcal{P}$ rather than free-form generation, ensuring that each reasoning step follows the evidence graph. The correct answer is rendered from $\mathcal{P}$, while distractors are programmatically sampled from alternative STSG elements that are visually plausible but evidence-inconsistent. Finally, grounding metadata $\mathcal{M}$ links all referenced entities to their original spatio-temporal locations, making every QA instance fully traceable.

\noindent\textbf{Traceable Verification.} Finally, we use the grounding metadata $\mathcal{M}$ to retrieve corresponding frames, clips, and object trajectories as visual evidence. A verifier VLLM jointly examines the visual evidence and generated $(Q, A)$ pair for factual consistency. Samples with inconsistent reasoning or incorrect grounding are filtered, ensuring the final dataset is evidence-aligned and free from hallucinated annotations. 

\section{The V-STORM Dataset}

To systematically investigate the performance of MLLMs in fine-grained spatiotemporal localization and reasoning tasks, we constructed the V-STORM Dataset. This dataset comprises two distinct components: a rigorously human-annotated Benchmark for evaluation, and a large-scale Training Split constructed for model training. While distinguished by their annotation precision and construction methodologies, both splits share a unified data structure and task definition, thereby supporting large-scale training and rigorous evaluation within a single consistent paradigm. Examples of tasks are provided in the Supplementary Material, Sec. 5.

\subsection{Scope \& Coverage}

V-STORM is built from several public large-scale video datasets, including SAV~\cite{ravi2024sam}, ActivityNet~\citep{yu2019activitynet}, Panda~\citep{wang2020panda}, WebVid~\citep{bain2021frozen}, and InternVid~\citep{wang2023internvid}. A unified sampling frame rate and a maximum pixel budget are applied to ensure consistent perceptual granularity and manageable inference.

\begin{figure}[t] 
  \centering
  \includegraphics[width=0.5\textwidth]{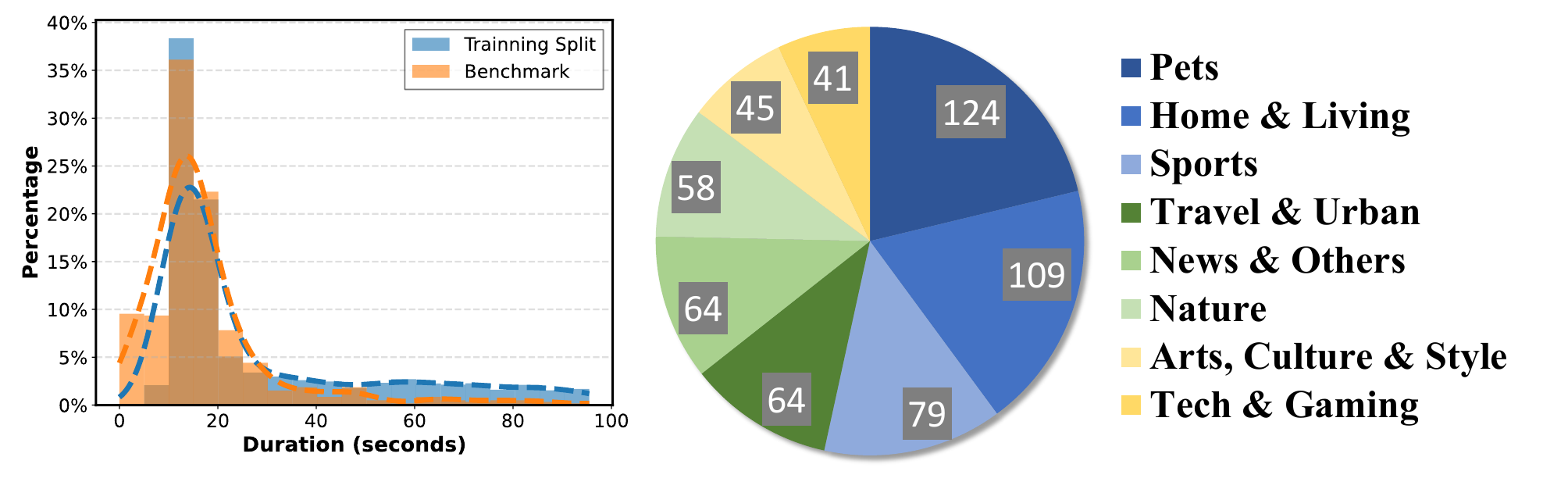} 
  \caption{Distribution of video lengths in the Training Split and Benchmark, and video topics in the Benchmark.}
  \label{fig:data1}
\end{figure}

As shown in Fig.~\ref{fig:data1}, the V-STORM dataset spans both short and medium-length videos, supporting fine-grained spatiotemporal localization while preserving enough context for event-level reasoning. The V-STORM dataset covers diverse real-world domains such as sports, traffic, cooking, animals, natural scenery, home life, and music, providing a rich semantic foundation for studying fine-grained video reasoning that demands precise spatio-temporal perception and cross-temporal reasoning.

\subsection{Capability Taxonomy \& Case Study}

To facilitate fine-grained spatio-temporal reasoning, V-STORM provides Grounded Chain-of-Thought (GCoT) annotations that embed spatio-temporal grounding into each reasoning step. As Fig.~\ref{fig:intro}(c) illustrates, GCoT adopts a step-by-step natural language format, with every intermediate step tied to concrete temporal segments and spatial regions. This design guides models to explicitly utilize fine-grained visual evidence during multi-step reasoning, rather than relying solely on abstract linguistic patterns.

In GCoT, the temporal and spatial references are not generated freely but are derived from the spatio-temporal annotations defined during data construction. By coupling each reasoning step with explicit grounding metadata, the reasoning process becomes directly traceable to verifiable visual evidence. This formulation encourages models to not only produce logically coherent explanations, but also to perform reasoning that is consistently anchored in correct temporal and spatial locations.

\subsection{Data Verification \& Quality Control}

V-STORM employs rigorous quality control to ensure reliability. For the Benchmark split, eight trained experts manually inspected every sample, verifying question formulation, answer correctness, and spatio-temporal consistency. For the Training split, we used metadata-guided automated verification (Fig. \ref{fig:pipeline} stage 3), supplemented by stratified human auditing. As Table~\ref{tab:human_audit_resize} shows, this hybrid strategy achieves stable logical and annotation accuracy across categories, ensuring high-quality scaling without exhaustive manual effort.

\begin{table}[t]
\centering
\small
\setlength{\tabcolsep}{1mm}
\caption{Human audit results on the training split.}
\label{tab:human_audit_resize}
\begin{tabular}{lcccc}
\toprule
\textbf{Question Category} & \textbf{Samples} & \textbf{Before} & \textbf{After} & \textbf{Imp. ($\Delta$)} \\
\midrule
Frame-Level Perception & 100 & 82\% & 98\% & +16\% \\
Segment-Level Perception & 100 & 78\% & 96\% & +18\% \\
Temporal Reasoning & 100 & 71\% & 92\% & +21\% \\
Compositional Reasoning & 100 & 64\% & 94\% & +30\% \\
\midrule
\textbf{Overall / Average} & \textbf{400} & \textbf{73.8\%} & \textbf{95.0\%} & \textbf{+21.2\%} \\
\bottomrule
\end{tabular}
\end{table}

\begin{table*}[!t] 
\centering
\small
\caption{\textbf{Performance of representative commercial and open-source multimodal models on the V-STORM Benchmark}. Results are reported across four capability categories using accuracy \textbf{(Acc)} as the primary metric, with auxiliary Reasoning \textbf{(R)} and Grounding \textbf{(G)} scores for evidence-aligned spatio-temporal reasoning. Best scores are \textbf{bold}, and second-best are \underline{underlined}.}
\label{tab:main_results}

\setlength{\tabcolsep}{1mm}
\renewcommand{\arraystretch}{1.0}

\begin{tabular}{l c c c c c c c c c c c c}
\toprule
\textbf{Model} & \textbf{FPS} & 
\textbf{Frame-Level} & 
\textbf{Segment-Level} & 
\multicolumn{3}{c}{\textbf{Temporal Reasoning}} & 
\multicolumn{3}{c}{\textbf{Comp. Reasoning}} & 
\multicolumn{3}{c}{\textbf{Overall}} \\

\cmidrule(lr){5-7} \cmidrule(lr){8-10} \cmidrule(lr){11-13}
& & \textbf{Perception (Acc)} & \textbf{Perception (Acc)} & \textbf{Acc} & \textbf{R} & \textbf{G} & \textbf{Acc} & \textbf{R} & \textbf{G} & \textbf{Acc} & \textbf{R} & \textbf{G} \\
\midrule

 
\multicolumn{13}{c}{\textit{\textbf{Commercial Models}}} \\ 
\midrule

    GPT-5.2        & 2 & \underline{75.95} & 90.36 & 79.12 & 3.25 & 3.58 & \underline{41.71} & 2.63 & 3.75 & \underline{69.91} & 2.80 & 3.70 \\
Qwen3.6-plus    & 2 & 73.42 & 91.30 & \textbf{80.93} & \textbf{3.43} & \textbf{3.97} & 40.36 & 2.75 & 3.96 & 69.38 & \underline{2.93} & 3.97 \\
Gemini3-Flash    & 1 & 69.71 & 87.81 & 75.52 & 3.02 & 3.61 & 34.84 & 2.51 & \underline{4.06} & 64.97 & 2.65 & 3.94 \\
\midrule


\multicolumn{13}{c}{\textit{\textbf{Open-Source Models}}} \\
\midrule
Qwen3-VL-2B-Instruct  & 2  & 58.99 & 77.41 & 63.66 & 2.31 & 3.07 & 19.85 & 1.64 & 3.14 & 52.83 & 1.82 & 3.22 \\
Qwen3-VL-4B-Instruct  & 2 & 65.44 & 90.64 & 79.64 & 3.18 & 3.67 & 27.00 & 2.16 & 3.82 & 62.81 & 2.44 & 3.78 \\
InternVL3.5-4B-Instruct   & 2 & 59.49 & 84.22 & 68.56 & 3.01 & 2.80 & 23.80 & \underline{3.27} & 3.65 & 56.99 & 2.06 & 3.43 \\
VideoLLaMA3-7B & 2 & 46.20 & 72.50 & 58.51 & 0.69 & 1.48 & 15.99 & 0.66 & 1.38 & 46.35 & 0.67 & 1.40 \\
Qwen3-VL-8B-Instruct & 2 & 68.48 & \underline{91.40} & 78.87 & 2.82 & 2.69 & 27.38 & 2.33 & 3.86 & 63.81 & 2.20 & 3.55 \\
Qwen3-VL-32B-Instruct  & 2 & 71.27 & \textbf{92.82} & \underline{80.67} & \underline{3.36} & \underline{3.92} & 32.27 & 2.36 & \underline{4.06} & 66.72 & 2.63 & \underline{4.03} \\

\midrule
\rowcolor{ourshade}
Qwen3-VL-2BI+\textbf{{V-STORM}} 
& 2  
& \textbf{76.84} & 88.94 & 75.52 & 3.11 & 3.77 & \textbf{73.28} & \textbf{3.48} & \textbf{4.61} & \textbf{79.42} & \textbf{3.38} & \textbf{4.38} \\

\bottomrule
\end{tabular}
\end{table*}

\section{Experiments}
This section empirically evaluates V-STORM's design rationality and validates its effectiveness as both a unified benchmark and a training resource. Our experiments address three questions: (i) whether V-STORM effectively discriminates MLLMs' fine-grained spatio-temporal perception and reasoning; (ii) the performance characteristics of mainstream models on this benchmark; and (iii) whether V-STORM's training split provides effective supervision for model training. Notably, this section introduces no new model architectures or algorithms; all experiments solely verify the dataset's diagnostic value and utility.

\subsection{Evaluation Setup}

\noindent\textbf{Evaluation Protocol.}
The V-STORM benchmark adopts a unified multiple-choice QA format. Besides selecting the correct answer, our prompt explicitly instructs models to generate a Grounded Chain-of-Thought (GCoT) by citing fine-grained spatio-temporal evidence, including relevant timestamps and object locations, for each reasoning step. This protocol encourages evidence-grounded reasoning rather than answer-only prediction, enabling joint evaluation of answer correctness, reasoning quality, and grounding faithfulness. To ensure fair comparison, all models are evaluated under a unified zero-shot prompting protocol with a standardized output format for automatic parsing and scoring. Abstract prompt templates are provided in Fig. \ref{fig:failure} and the full version is provided in Supplementary Material, Sec. 4.1.

\noindent\textbf{Evaluation Metrics.}
We evaluate models using three complementary metrics: Answer Accuracy, Reasoning Score, and Grounding Score. Accuracy measures task completion; Reasoning Score assesses logical consistency of generated reasoning chains via Qwen; Grounding Score evaluates alignment of cited evidence with annotated timestamps and object locations. These metrics disentangle answer correctness, reasoning quality, and evidence localization.
To validate our scoring pipeline, we conduct a preliminary agreement study on a sample set by comparing Qwen with ChatGPT and Gemini.
All three models show high inter-model agreement with a quadratic weighted kappa of 0.8841. We further verify the scoring quality via human expert review. Detailed metric definitions are in the Supplementary Material, Sec. 4.2.

\begin{figure}[t]
\centering
\includegraphics[width=1\linewidth]{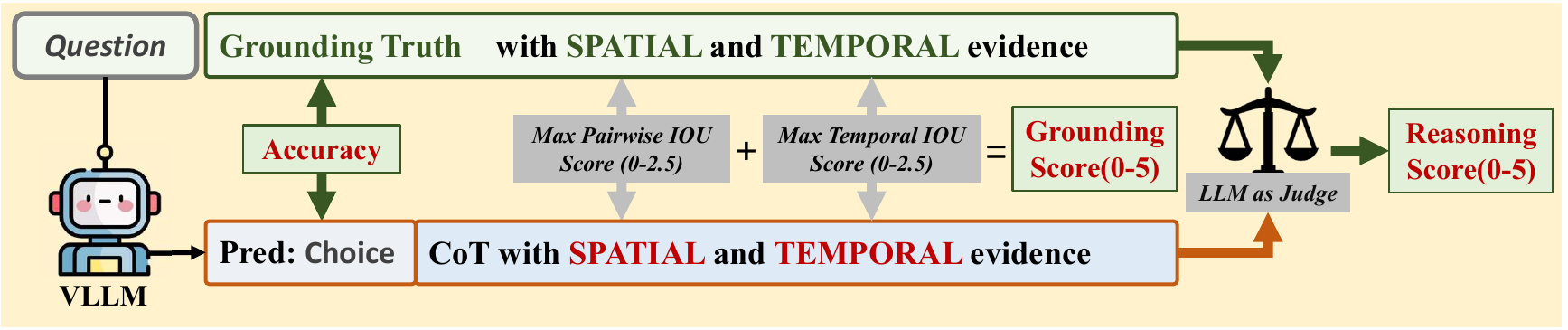}
\caption{
Definition of Answer Accuracy, Reasoning Score, and Grounding Score.
}
\label{fig:failure}
\end{figure}

\begin{table}[ht]
\centering
\caption{Human audit accuracy (Acc) across STSG construction stages: node category/bbox, edge relation, segment trajectory/event, and video-level identity/event integration.}
\small
\setlength{\tabcolsep}{1mm}
\begin{tabular}{@{}lcccc@{}}
\toprule
\textbf{Metric} & \textbf{FSG-Node} & \textbf{FSG-Edge} & \textbf{Segment STSG} & \textbf{Video STSG} \\ \midrule
\textbf{Acc}
&  92\%     & 88\%     & 91\%               & 95\%             \\ \bottomrule
\end{tabular}
\label{tab:human_audit_stsg}
\end{table}

\noindent\textbf{Test Settings.}
We evaluate representative open-source and proprietary multimodal video models, including the Qwen3-VL~\cite{bai2025qwen3}, GPT~\cite{hurst2024gpt}, Gemini~\cite{team2023gemini}, VideoLLaMA3~\cite{zhang2025videollama}, and InternVL~\cite{wang2025internvl3} series. To ensure fair comparison, all models are evaluated under a unified maximum pixel budget, video sampling rate (FPS = 2), and prompt template, except Gemini-3-Flash, which uses its default setting (FPS = 1). Open-source models use greedy decoding (temperature = 0) with a maximum generation length of 2048 tokens.

\subsection{Pipeline Reliability Validation} 
We further evaluate whether errors accumulate during the hierarchical STSG construction process through stage-wise human auditing. As shown in Table~\ref{tab:human_audit_stsg}, high audited accuracies are consistently maintained across frame-level node grounding (92\%), relation construction (88\%), segment-level aggregation (91\%), and video-level integration (95\%), suggesting that the proposed pipeline preserves reliable spatio-temporal grounding throughout successive construction stages. In addition, the metadata-guided verification strategy improves the audited accuracy of generated training samples from 73.8\% to 95.0\% (Table~\ref{tab:human_audit_resize}), indicating that noisy or inconsistent samples are effectively filtered before inclusion in the final training set. Together, these results demonstrate that V-STORM provides a reliable and traceable pipeline for large-scale evidence-grounded data construction.
\subsection{Benchmark Results of Existing Models}

Table~\ref{tab:main_results} summarizes the performance of representative MLLMs on the V-STORM Benchmark. Existing MLLMs achieve strong performance on coarse semantic perception, but their accuracy consistently declines as tasks require fine-grained temporal localization or compositional reasoning. Although leading commercial models outperform open-source counterparts overall, all models exhibit substantial performance drops on tasks requiring precise spatio-temporal understanding, indicating that current MLLMs still struggle with fine-grained grounded video reasoning.

A closer examination of the auxiliary metrics further reveals that model failures vary with reasoning complexity. For Temporal Reasoning tasks, high Answer Accuracy is consistently accompanied by high Grounding scores, suggesting that current MLLMs can reliably identify the required spatio-temporal evidence for relatively simple reasoning. However, in Compositional Reasoning, Answer Accuracy drops markedly despite maintaining strong Grounding performance. This observation indicates that current models are often able to locate the correct evidence, yet still struggle to integrate multiple grounded observations into coherent reasoning chains. These results demonstrate that V-STORM effectively separates evidence localization from reasoning ability, enabling diagnosis of failure modes that remain hidden under conventional answer-only evaluation.

\subsection{Effectiveness of V-STORM-260K}
To evaluate V-STORM-260K as a supervision resource, we fine-tune Qwen3-VL-2B-Instruct and compare it with representative commercial and open-source MLLMs on the V-STORM Benchmark. As shown in Table 3, V-STORM fine-tuning substantially improves fine-grained spatio-temporal reasoning. The resulting 2B model achieves an Overall Accuracy of \textbf{79.42}, outperforming all evaluated models despite its substantially smaller scale. Compositional Reasoning accuracy increases from 19.85 to \textbf{73.28}, indicating strong task-relevant supervision for multi-step reasoning. The Grounding Score improves from 3.22 to 4.38, the highest among evaluated models, demonstrating improved alignment between predictions and supporting spatio-temporal evidence. These results demonstrate that V-STORM-260K effectively improves both answer accuracy and evidence grounding, with particularly substantial gains in compositional reasoning.

Fig.~\ref{fig:failure} presents representative failure cases of existing models on V-STORM. 
In the top case, the model identifies the correct subject and produces a region covering the target, but fails to capture the required semantic granularity by selecting the full body instead of the face. 
In the bottom case, the model successfully grounds the object in the initial scene but loses its spatio-temporal correspondence after a temporal offset, leading to incorrect tracking and reasoning. 
These examples demonstrate that current MLLMs may achieve plausible semantic understanding while still struggling with fine-grained evidence alignment and temporal consistency.

\begin{figure}[t]
\centering
\includegraphics[width=1\linewidth]{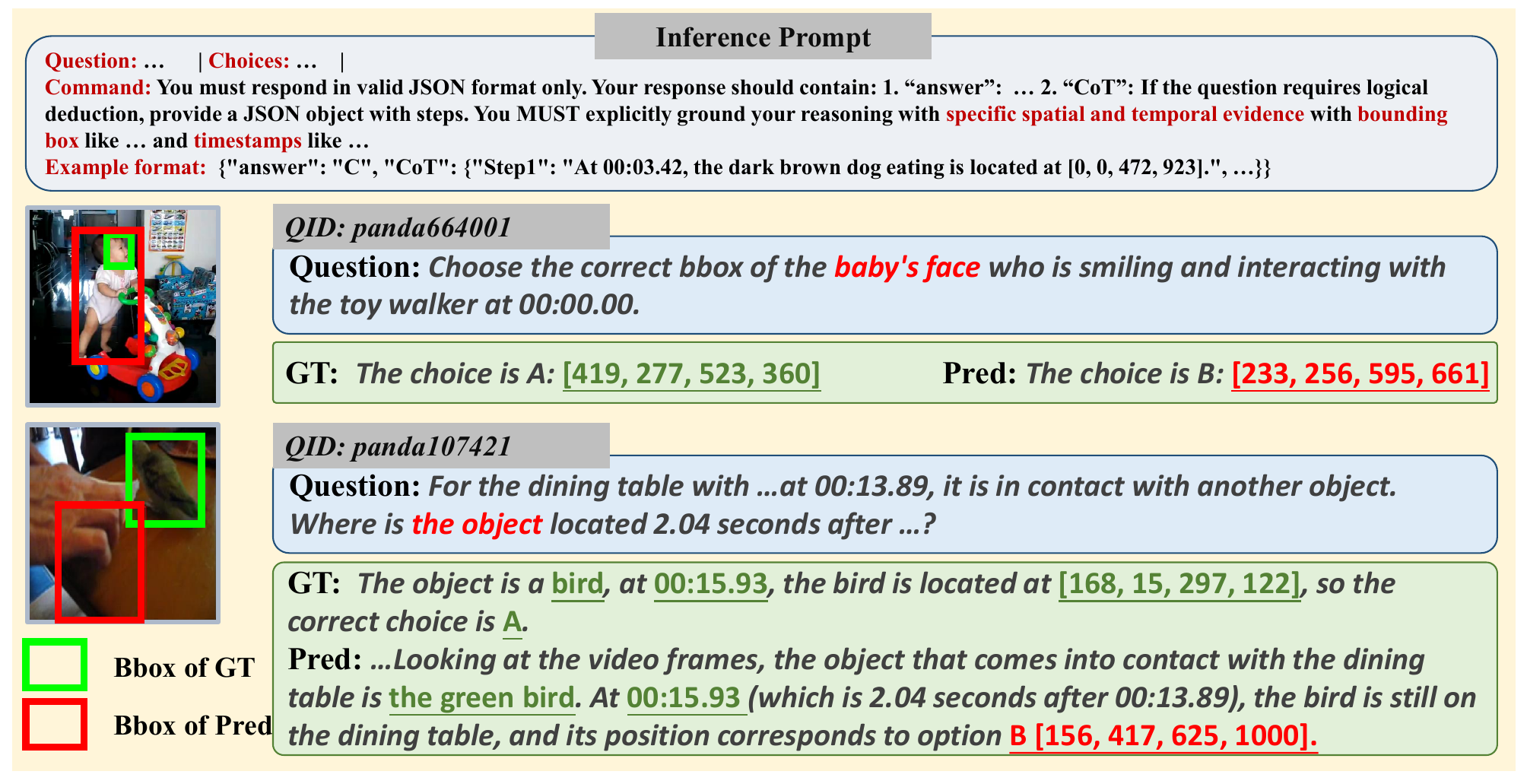}
\caption{
Inference Prompt (Top) and Failure Cases on the V-STORM benchmark.
Middle: the model recognizes the correct subject but selects a wrong region due to semantic granularity mismatching.
Bottom: the model correctly grounds the object at first but fails to track it later.
}
\label{fig:failure}
\end{figure}

\subsection{Robustness and Generalization Analysis}

We further evaluate our approach from two complementary perspectives: generalization to out-of-distribution (OOD) settings and the effects of different reasoning supervision strategies, providing additional insights into model behavior under varying conditions.

\noindent\textbf{OOD Generalization.}
We further evaluate the generalization capability of models trained on V-STORM under out-of-distribution (OOD) settings. As shown in Table~\ref{tab:ood}, models trained with our V-STORM-260K maintain stable performance across diverse evaluation distributions, indicating strong generalization beyond the training domain.
V-STORM yields consistent improvements across representative benchmarks, particularly those requiring fine-grained spatio-temporal understanding and reasoning. Notably, we observe gains on VideoRefer-Bench~\cite{yuan2025videorefer} and NExT-QA~\cite{xiao2021next}, which emphasize spatio-temporal grounding, as well as on EOC-Bench~\cite{yuan2025eoc}, which focuses on object-centric temporal modeling. In addition, performance improvements on the reasoning-oriented MMBench-Video~\cite{fang2024mmbench} further demonstrate the effectiveness of V-STORM.

\begin{table}[t]
\centering
\small
\setlength{\tabcolsep}{3pt}
\caption{Performance on OOD benchmarks.}
\begin{tabular}{lcccc}
\toprule
\textbf{Method} & \textbf{VideoRefer} & \textbf{NExT-QA} & \textbf{MMBench-V} & \textbf{EOC} \\
\midrule
Qwen3-VL-2B & 67.20 & 74.88 & 1.3078 & 40.82 \\
\rowcolor{ourshade}\textbf{+ VSTORM} & \textbf{68.60} & \textbf{75.92} & \textbf{1.3163} & \textbf{42.32} \\
\bottomrule
\end{tabular}
\label{tab:ood}
\end{table}

\begin{table}[t]
\centering
\small
\caption{Impact of different reasoning supervision strategies on various spatio-temporal tasks. (FG: Fine-grained Task, CG: Coarse-grained Task, CCoT: Coarse-grained CoT)}
\setlength{\tabcolsep}{1mm}
\begin{tabular}{lccccc}
\toprule
\textbf{Strategy} 
& \textbf{FG} 
& \textbf{CG} 
& \textbf{General} 
& \textbf{Complex} 
& \textbf{Overall} \\
\midrule
baseline 
& 60.25 & 86.77 & 73.91 & 22.11 & 58.08 \\

sft w/o CoT (95k) 
& 72.66 & 87.52 & 72.94 & 70.84 & 76.87\\

sft w/ CCoT (95k) 
& 74.43 & 87.43 & 74.23 & 71.21 & 77.57 \\

sft w/ GCoT (95k) 
& \textbf{76.20} & \textbf{87.71} & \textbf{75.26} & \textbf{73.19} & \textbf{78.84} \\
\midrule

GCoT vs Non-CoT 
& \textbf{\textcolor{green!70!black}{+3.54}}
& \textbf{\textcolor{darkgray}{+0.19}}
& \textbf{\textcolor{green!70!black}{+2.32}}
& \textbf{\textcolor{green!70!black}{+2.35}}
& \textbf{\textcolor{green!70!black}{+1.97}}  \\

GCoT vs CCoT 
& \textbf{\textcolor{green!70!black}{+1.77}} 
& \textbf{\textcolor{darkgray}{+0.28}}
& \textbf{\textcolor{green!70!black}{+1.03}} 
& \textbf{\textcolor{green!70!black}{+1.98}}
& \textbf{\textcolor{green!70!black}{+1.27}} \\

\bottomrule
\end{tabular}
\label{tab:ablation}
\end{table}

\noindent\textbf{Impact of Reasoning Supervision.}
We further investigate different reasoning supervision strategies, including \textbf{coarse-grained CoT (CCoT)} and our \textbf{GCoT}, using the same V-STORM subset ($\sim$95k) and Qwen3-VL-2B-Instruct as the baseline model. As shown in Table~\ref{tab:ablation}, SFT without explicit reasoning supervision already substantially improves performance, particularly on complex reasoning tasks (22.11$\rightarrow$70.84), demonstrating the value of task-specific spatio-temporal supervision. Both CCoT and GCoT provide additional improvements, suggesting that explicit reasoning supervision further benefits model learning. Notably, GCoT consistently achieves the best performance across all evaluated metrics, outperforming SFT without CoT by 1.97 points in Overall Accuracy and CCoT by 1.27 points. The gains are particularly pronounced on fine-grained perception (+3.54 over SFT without CoT) and complex reasoning (+2.35 and +1.98 over SFT without CoT and CCoT, respectively). These results suggest that \textbf{fine-grained spatio-temporal grounding provides more effective reasoning supervision than coarse-grained reasoning traces}, while the relatively modest improvement on coarse-grained tasks indicates that structured reasoning supervision is particularly beneficial for more demanding fine-grained and compositional reasoning.

\section{Conclusion}

In this paper, we present V-STORM, a structure-aware framework for scalable construction of fine-grained video reasoning data. By leveraging Spatio-Temporal Scene Graphs (STSGs) as a unified intermediate representation, V-STORM supports graph-grounded question generation with verifiable spatio-temporal evidence. Based on this framework, we construct a training set and a human-verified benchmark for evaluating fine-grained perception, evidence localization, and compositional reasoning. Our results demonstrate the value of graph-grounded supervision for learning evidence-aware video reasoning while enabling systematic evaluation of reasoning capabilities. Experiments reveal that although current MLLMs can accurately localize relevant evidence, composing multiple grounded observations into correct reasoning remains a major challenge.
We hope V-STORM serves as a valuable benchmark and training resource for future research on fine-grained video reasoning.

\bibliography{aaai2027}


\end{document}